\documentclass[pdflatex,sn-mathphys-num]{sn-jnl}% Math and Physical Sciences Numbered Reference Style
\usepackage{graphicx}%
\usepackage{multirow}%
\usepackage{amsmath,amssymb,amsfonts}%
\usepackage{amsthm}%
\usepackage{mathrsfs}%
\usepackage[title]{appendix}%
\usepackage{xcolor}%
\usepackage{textcomp}%
\usepackage{manyfoot}%
\usepackage{booktabs}%
\usepackage{algorithm}%
\usepackage{algorithmicx}%
\usepackage{algpseudocode}%
\usepackage{listings}%

\usepackage{silence}
\usepackage[switch]{lineno}

\begin{document}

\title[Article Title]{Sliding Gaussian ball adaptive growth: point cloud-based iterative algorithm for large-scale 3D photoacoustic imaging}

\author[1]{\fnm{Shuang} \sur{Li}}\email{jaeger\_ls@stu.pku.edu.cn}
\equalcont{These authors contributed equally to this work.}

\author[1]{\fnm{Yibing} \sur{Wang}}\email{ddffwyb@pku.edu.cn}
\equalcont{These authors contributed equally to this work.}

\author[2]{\fnm{Jian} \sur{Gao}}\email{jian\_gao@smail.nju.edu.cn}
\equalcont{These authors contributed equally to this work.}

\author[3]{\fnm{Chulhong} \sur{Kim}}\email{chulhong@postech.edu}

\author[3]{\fnm{Seongwook} \sur{Choi}}\email{swchoi715@postech.ac.kr}

\author[1]{\fnm{Yu} \sur{Zhang}}\email{zyuaiyi1\_@stu.pku.edu.cn}

\author[1]{\fnm{Qian} \sur{Chen}}\email{chen\_qian@stu.pku.edu.cn}

\author*[2]{\fnm{Yao} \sur{Yao}}\email{yaoyao@nju.edu.cn}

\author*[1, 4]{\fnm{Changhui} \sur{Li}}\email{chli@pku.edu.cn}

\affil*[1]{\orgdiv{Department of Biomedical Engineering, College of Future Technology}, \orgname{Peking University}, \city{Beijing}, \country{China}}

\affil*[2]{\orgdiv{School of Intelligence Science and Technology}, \orgname{Nanjing University}, \city{Suzhou}, \country{China}}

\affil[3]{\orgdiv{Department of Electrical Engineering, Convergence IT Engineering, Mechanical Engineering, and Medical Science and Engineering, Medical Device Innovation Center}, \orgname{Pohang University of Science and Technology}, \city{Pohang}, \country{Republic of Korea}}

\affil*[4]{\orgdiv{National Biomedical Imaging Center}, \orgname{Peking University}, \city{Beijing}, \country{China}}

\abstract{Large-scale 3D photoacoustic (PA) imaging has become increasingly important for both clinical and pre-clinical applications. Limited by cost and system complexity, only systems with sparsely-distributed sensors can be widely implemented, which desires advanced reconstruction algorithms to reduce artifacts. However, high computing memory and time consumption of traditional iterative reconstruction (IR) algorithms is practically unacceptable for large-scale 3D PA imaging. Here, we propose a point cloud-based IR algorithm that reduces memory consumption by several orders, wherein the 3D PA scene is modeled as a series of Gaussian-distributed spherical sources stored in form of point cloud. During the IR process, not only are properties of each Gaussian source, including its peak intensity (initial pressure value), standard deviation (size) and mean (position) continuously optimized, but also each Gaussian source itself adaptively undergoes destroying, splitting, and duplication along the gradient direction. This method, named the sliding Gaussian ball adaptive growth (SlingBAG) algorithm, enables high-quality large-scale 3D PA reconstruction with fast iteration and extremely low memory usage. We validated SlingBAG algorithm in both simulation study and in vivo animal experiments. The source code and data for SlingBAG, along with supplementary materials and demonstration videos, are now available in the following GitHub repository: \href{https://github.com/JaegerCQ/SlingBAG}{https://github.com/JaegerCQ/SlingBAG}.}

\keywords{Iterative 3D PA reconstruction, Point cloud-based model, Gaussian ball representation, Sparse sensor distribution}

\maketitle

\section{Introduction}\label{sec1}
Photoacoustic imaging (PAI), which uniquely combines ultrasound detection and optical absorption contrast, is the only non-invasive optical imaging method that can image centimeters deep in living tissue with superior spatial resolution compared to other optical imaging techniques. PAI has been widely employed in various biomedical studies, including both pre-clinical studies and clinical implementations~\cite{park2024clinical, wang2012photoacoustic, assi2023review, dean2017advanced, lin2022emerging, ntziachristos2024addressing}.

In recent years, significant progress has been made in the development of 3D PAI through the use of two-dimensional (2D) matrix arrays, demonstrating immense potential for in vivo applications, such as human peripheral vessel imaging~\cite{wray2019photoacoustic, li2024photoacoustic}, breast imaging~\cite{wang2021functional, han2021three}, and small animal imaging~\cite{sun2024real}. Various array configurations have been explored, ranging from spherical and planar arrays~\cite{matsumoto2018label, matsumoto2018visualising, ivankovic2019real, dean2013portable, nagae2018real, kim2023wide, piras2009photoacoustic, heijblom2012visualizing} to synthetic planar arrays that rely on scanning~\cite{li2024photoacoustic, wang2024comprehensive}. However, limited by cost and technical difficulties, it is a great challenge to fabricate a matrix array that satisfies the spatial Nyquist sampling requirement for real-time large-scale 3D PAI, which typically requires tens of thousands or more ultrasound transducer elements, as well as an equal number of parallel data acquisition channels. Therefore, real-time 3D PAI systems only have sparsely distributed sensors in reality, while it is well known that traditional back-projection image reconstruction methods will cause severe artifacts and a low signal-to-noise ratio (SNR) for extremely sparse arrays.

To address these challenges, researchers have employed iterative reconstruction (IR) methods to improve image quality with sparse sensor configurations. One of the first IR approaches to PAI was presented by Paltauf et al.~\cite{paltauf2002iterative}, wherein the differences between observed and simulated projections were iteratively minimized to improve image quality. Subsequent IR methods have incorporated various physical imaging factors. Then, De\'an-Ben et al.~\cite{dean2012accurate} developed an accurate model-based 3D reconstruction algorithm utilizing a sparse matrix model to calculate the theoretical pressure with reduced artifacts compared to back-projection methods. Wang et al.~\cite{wang2012investigation, wang2014discrete} demonstrated that iterative penalized least-squares algorithms, based on discrete-to-discrete (D-D) imaging models using expansion functions in several types of fixed spatial grids, with either quadratic smoothing or total variation (TV) norm penalties, could enhance the performance of small animal 3D PAI systems. Similarly, Huang et al.~\cite{huang2013full} introduced forward and backward operators based on k-space pseudospectral methods. Although those IR techniques have demonstrated great performance, for large-scale 3D PAI — where the spatial grid and system matrix scale is substantially increased, the computational load and memory demands make these methods less feasible~\cite{zhu2023mitigating}.

Researchers continued to explore new ways. Efficient numerical implementations of adjoint operators for PAI reconstruction were introduced by Arridge et al.~\cite{arridge2016adjoint}, who applied compressed sensing via Bregman iteration to reduce the required number of sensors. However, their approach was computationally intensive, limited to voxel counts on the order of $10^6$. Shang et al.~\cite{shang2019sparsity} approached the problem by creating a forward model using directly measured graphite point sources, but optimization inefficiencies and inaccuracies in the forward model hindered its practical application.

Besides traditional IR, deep learning approaches for PAI reconstruction ~\cite{hauptmann2018model, hauptmann2020deep, zheng2023deep} offer improvements in computational efficiency and image quality. However, they heavily rely on large, specialized training datasets, which can be difficult to obtain, and lack generalizability across different imaging systems or under varying conditions. This limits their broader applicability, particularly in diverse clinical settings where imaging conditions are not consistent.

Recently, Huynh et al.~\cite{huynh2024fast} employed an iterative model-based reconstruction algorithm to achieve 3D PAI. Their PAI system is based on a small transparent Fabry-Perot flat scanner, which can be closely attached to the local flat tissue surface, substantially reducing the computing spatial space. This approach is not suitable for general large-scale 3D PAI for targets with uneven outer surfaces (like human thyroid tumor or whole breast tissue). Besides, they still need seconds to scan optical US sensors for one image acquisition, not a real-time PAI system.

Similar challenge of reconstructing 3D images from sparse-view also exists in the computer vision field. To tackle this challenge, differentiable rendering technologies such as NeRF~\cite{mildenhall2021nerf} and 3D Gaussian Splatting (3DGS)~\cite{kerbl20233Dgaussians} are developed. Differentiable rendering technology~\cite{weiss2021differentiable, yariv2021volume, barron2021mip, chen2022tensorf, yao2022neilf} enables self-supervised reconstruction of 3D scenes directly from scene observations, usually multi-view images. The photometric difference between the rendered image generated by the differentiable rendering pipeline and the actual captured image is utilized as a loss function, guiding the reconstruction of the entire scene through gradient back-propagation. NeRF, for instance, uses multi-layer perceptrons (MLPs) to map 3D spatial positions to colors and densities, while newer methods like InstantNGP use hash grids to store features at different spatial scales, allowing for faster and more memory-efficient reconstructions~\cite{muller2022instant}. Those pioneering works inspire new IR methods for PAI. A recent work reported NeRF-based paradigms to perform 2D PAI image reconstruction~\cite{yao2024sparse}. However, that method relies on dense sampling, which makes it computationally prohibitive for large-scale 3D PAI.

In contrast to NeRF, the 3D Gaussian Splatting (3DGS)~\cite{kerbl20233d} effectively addresses these computational challenges in 3D computer graphics. Based on point cloud instead of spatial grid, 3DGS applies object-order rendering by representing scenes as Gaussian ellipsoids, which are projected and composited onto a 2D screen using Elliptical Weighted Average (EWA)~\cite{zwicker2002ewa} for alpha blending. As a differentiable rendering strategy using explicit scene representation, 3DGS enables efficient, high-quality 3D reconstruction and real-time rendering.

Inspired by 3DGS, we propose the sliding Gaussian ball adaptive growth algorithm (SlingBAG) for 3D PAI reconstruction. Distinct from traditional voxel grids, this method stores information of PA source in the form of point cloud that undergoes iterative optimization, and the 3D PA scene is modeled as a series of Gaussian-distributed spherical sources. In order to fulfill the differentiable condition, we constructed a novel ``differentiable rapid radiator" model based on the fact that PA spherical source has the analytical solution for the wave equation in acoustically non-viscous homogeneous medium. Each point cloud PA source is composed of multiple concentric spheres, with each sphere’s radius and initial PA pressure determined by the discretized Gaussian distribution. PA signals are the superposition of waves from spatially distributed Gaussian-distributed sources with specific peak intensity (pressure value), standard deviation (size), and mean value (spatial position). By minimizing the discrepancies between the predicted and the actual PA signals, we iteratively refine the point cloud and ultimately realize the 3D PAI reconstruction by converting the reconstruction result from point cloud into voxel grid.

This work introduces the SlingBAG algorithm, detailing its differentiable rapid radiator, point cloud optimization, and physics-based point cloud-to-voxel shader. Validated by both simulation and in vivo experiments, the proposed SlingBAG algorithm enables high-quality large-scale 3D PAI for vast potential applications with sparse sensor distributions.

\section{Results}\label{sec2}

\subsection{3D PA image reconstruction under extremely sparse planar array}\label{subsec2_1}

\begin{figure}[H]
\centering
\includegraphics[width=0.92\textwidth]{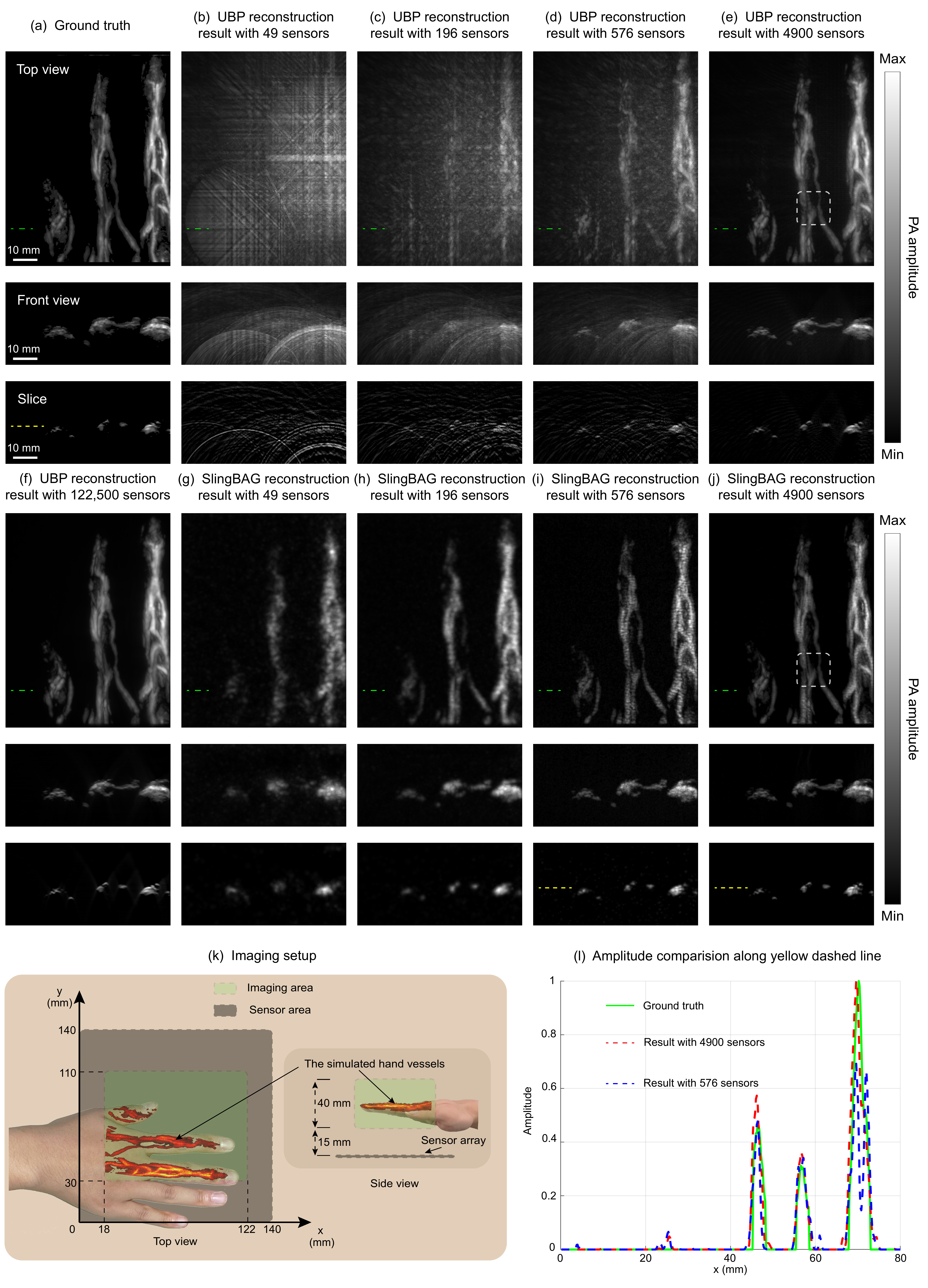}
\caption{Comparison of 3D photoacoustic reconstruction results under extremely sparse planar arrays. (a) Top-view maximum amplitude projection, front-view maximum amplitude projection, and the cross-section slice at the green dashed line marked in the top-view-MAP of the acoustic source. (b-f) Top-view maximum amplitude projection, front-view maximum amplitude projection, and the cross-section slice at the green dashed line marked in the top-view-MAP of the UBP reconstruction result with 49, 196, 576, 4900 and 122,500 sensors. (g-j) Top-view maximum amplitude projection, front-view maximum amplitude projection, and the cross-section slice at the green dashed line marked in the top-view-MAP of the SlingBAG reconstruction result with 49, 196, 576 and 4900 sensors. (k) Imaging setup. (l) Intensity distribution curves along the yellow dashed line in the vertical coordinate of the slices from the acoustic source, 576-sensor reconstruction, and 4900-sensor reconstruction. (Scale: 10 mm.)}\label{fig1}
\end{figure}

We first used simulated hand vessels as the target tissue to demonstrate the high performance of the SlingBAG algorithm. The simulated data is based on the 3D PA imaging results of human peripheral hand vessels from previous work~\cite{li2024photoacoustic}, which was used as the ground truth (Fig. \ref{fig1}(a)). At a grid point spacing of 0.2 mm, the total grid size of the imaging area is $400 \times 520\times 200$ ($80 \text{ mm}\times 104\text{ mm}\times 40\text{ mm}$). To simulate 3D PA imaging, the initial dense planar matrix array has 122,500 sensors ($350 \times 350$, with a pitch of 0.4 mm) uniformly placed on a plane covering $140 \text{ mm}\times 140\text{ mm}$ area, and the plane of the matrix array is about 15 mm below the imaging area (Fig. \ref{fig1}(k)).

For sparse detector arrangements, we undersampled the original 122,500 sensors into four different sparse matrix arrays with total sensor numbers of 49, 196, 576, and 4900, respectively. The pitches between adjacent sensors in these four different configurations are 20 mm, 10 mm, 6 mm, and 2 mm, respectively. Because current traditional IR methods cannot handle such large-scale cases, we compared the proposed SlingBAG algorithm with the widely-used universal back-projection (UBP) algorithm~\cite{xu2005universal}.

The reconstruction results are shown in Fig. \ref{fig1}(b-j). For each group of images, the three sub-images from top to bottom respectively show the Top-view maximum amplitude projection (MAP), Front-view MAP, and a cross-section slice along the green dashed line marked in the Top-view MAP of the 3D reconstruction result. More 3D reconstruction results using both algorithms with different array configurations can be found in Supplementary Video 1-3.

Under extremely sparse sensor configurations, the UBP reconstruction results have serious artifacts as shown in Fig. \ref{fig1}(b-d)), whereas the SlingBAG algorithm demonstrates remarkable 3D reconstruction capability (Fig. \ref{fig1}(g-i)); it even reconstructs a well-recognized outline of the hand vasculature with only 49 sensors. As the sensor matrix becomes denser, the quality of SlingBAG's reconstruction improves rapidly. With 576 sensors, SlingBAG already achieves high-quality reconstruction, while the UBP results still exhibit severe artifacts (Fig. \ref{fig1}(d)(i)). When with a total of 4900 sensors, the SlingBAG reconstruction (Fig. \ref{fig1}(j)) is nearly identical to the ground truth (Fig. \ref{fig1}(a)). In contrast, the UBP result (Fig. \ref{fig1}(e)) remains blurry due to reconstruction artifacts, making it difficult to discern fine vascular details (marked by the white dotted box). Notably, even when using all 122,500 sensors for UBP reconstruction (Fig. \ref{fig1}(f)), the result still cannot match the quality of SlingBAG's result with only 4900 sensors. This is because the finite size of the detection matrix array has the so-called ``limited view" problem. However, the SlingBAG algorithm performs exceptionally well in limited view conditions, which is another advantage of IR-based algorithms over back-projection methods.

In Fig. \ref{fig1}(l), we plotted the normalized PA value distribution along the yellow dashed line in the slice images for comparison (the dashed line's length does not represent the horizontal axis of the curve). The green solid line represents the ground truth, the red dashed line represents the reconstruction result from 4900 sensors, and the blue dashed line represents the reconstruction result from 576 sensors. As the number of sensors increases from 576 to 4900, the intensity distribution along the yellow dashed line in the slice progressively approximates the true value. The differences between the SlingBAG reconstruction results and the ground truth at several ``peaks" are also substantially diminished with 4900 sensors, indicating that the SlingBAG algorithm's reconstruction can achieve high quantitative accuracy with very sparse sensor distributions.

\begin{figure}[ht]
\centering
\includegraphics[width=0.97\textwidth]{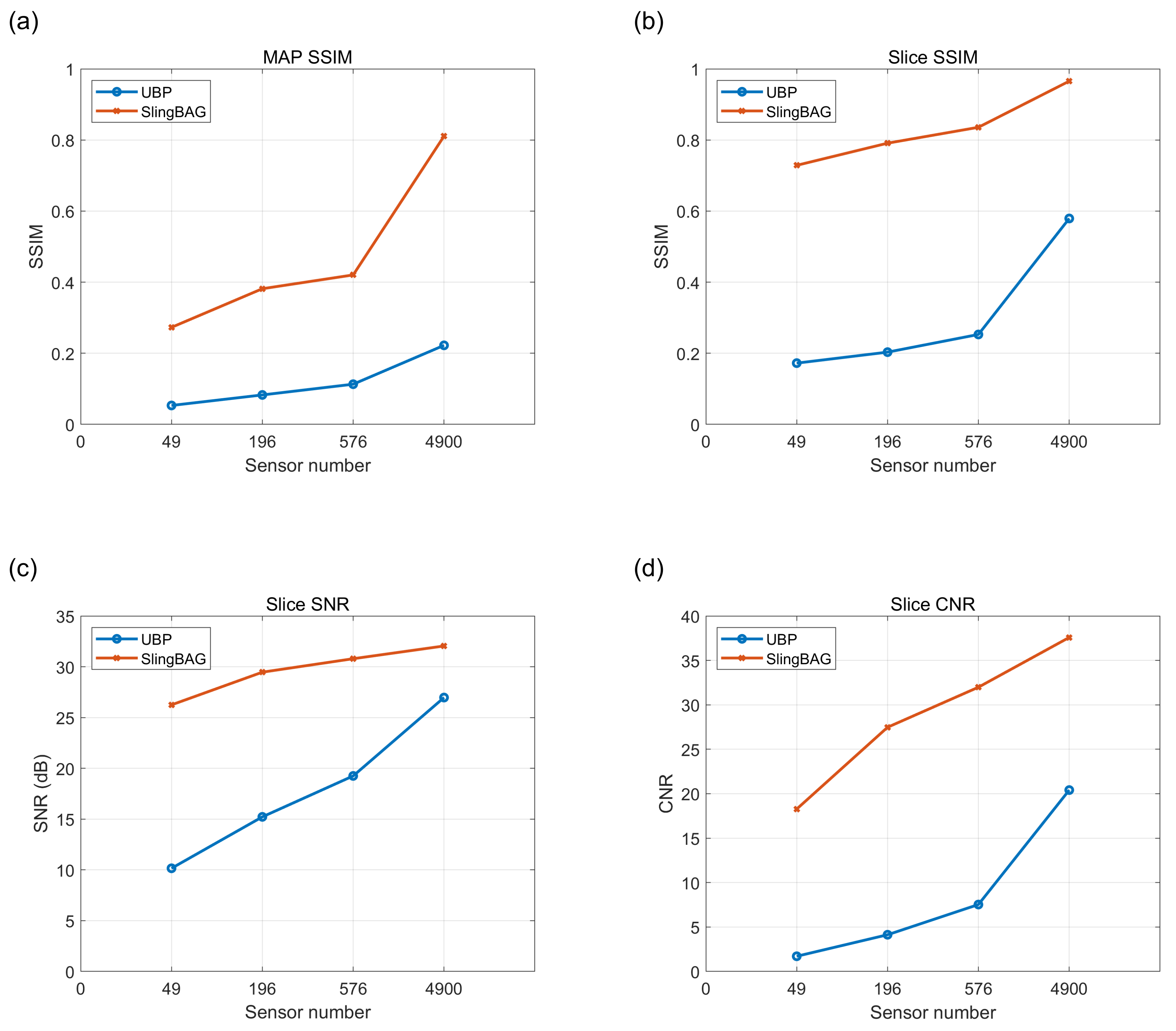}
\caption{Assessment for image quality of 3D photoacoustic reconstruction results under extremely sparse planar arrays. (a) Structural similarity index (SSIM) of the Top-view maximum amplitude projection of the reconstruction results. (b) Structural similarity index (SSIM) of the slice of the reconstruction results. (c) The signal-to-noise ratio (SNR) of the slice of the reconstruction results. (d) The contrast-to-noise ratio (CNR) of the slice of the reconstruction results.}\label{fig2}
\end{figure}

Next, to quantitatively assess the reconstruction results of the SlingBAG and UBP algorithms, we calculated the structural similarity index (SSIM) of the Top-view maximum amplitude projection (MAP), the SSIM of the slice, the signal-to-noise ratio (SNR), and the contrast-to-noise ratio (CNR) (Fig. \ref{fig2}) under different sensor configurations. In the SSIM for the Top-view MAP (Fig. \ref{fig2}(a)), for the result using 4900 sensors, the UBP algorithm achieves only a score of 0.22, whereas the SlingBAG result reaches 0.81. This directly reflects the high fidelity of the 3D reconstruction by SlingBAG. In Fig. \ref{fig2}(b), the SSIM for the slice similarly shows that under the same sensor configurations, SlingBAG has an unparalleled advantage in reconstructing vascular details, achieving an SSIM of 0.84 with just 576 sensors, while the UBP result under the same conditions falls below 0.3. When the number of sensors reaches 4900, the SSIM of SlingBAG for the slice hits 0.97, effectively indicating a high-accuracy reconstruction. Additionally, Fig. \ref{fig2}(c-d) shows that the SlingBAG reconstruction results exhibit significantly higher signal-to-noise ratio (SNR) and contrast-to-noise ratio (CNR) compared to the UBP results. In the case with 4900 sensors, the CNR of the SlingBAG result reaches approximately 40.

\begin{figure}[H]
\centering
\includegraphics[width=1.0\textwidth]{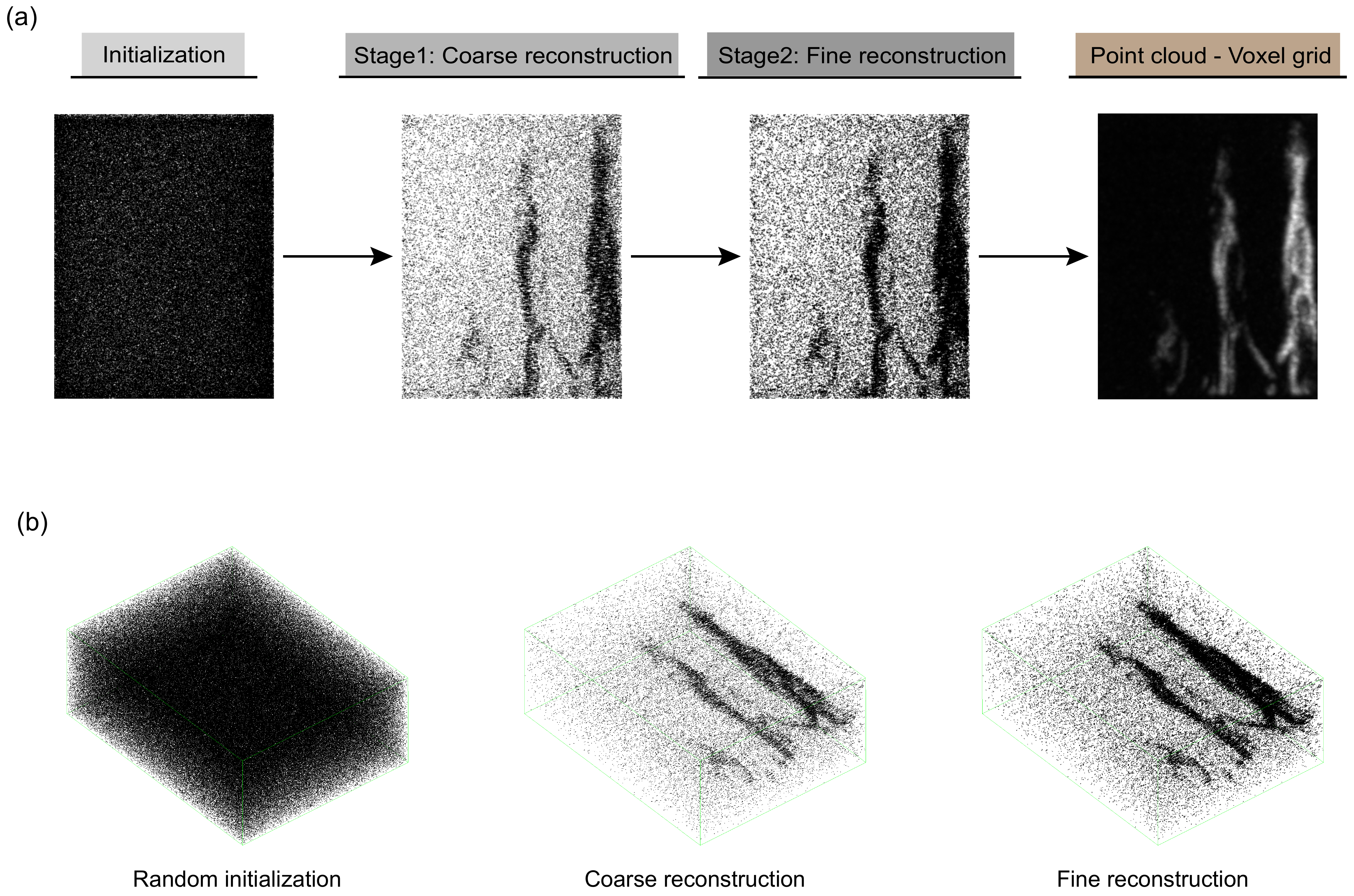}
\caption{Visualization of SlingBAG iterative results. (a) Visualization of initialization point cloud, coarse reconstruction point cloud, fine reconstruction point cloud and the final voxel grid reconstruction result. (b) 3D visualization of point clouds at various stages of the entire reconstruction process, including random initialization, coarse reconstruction stage, and fine reconstruction stage.}\label{fig3}
\end{figure}

We further visualized the iterative results of our reconstruction algorithm's point cloud for the 196-sensor array in Fig. \ref{fig3}. We first initialized a relatively dense point cloud distribution randomly within the reconstruction region. At the coarse reconstruction stage, both the position update and point duplication were disabled, only the peak initial pressure and standard deviations of each Gaussian ball underwent updates. Then, we got a preliminary convergence result in the coarse reconstruction stage. Next, we enabled the model's position update and point cloud duplication features. It can be seen that the converged shape of the three finger vessels in the coarse reconstruction gradually becomes denser, along with finer details showing up (Fig. \ref{fig3}(a)). Fig. \ref{fig3}(b) shows the 3D visualization results of the point cloud throughout the entire reconstruction process, from random initialization to the coarse reconstruction stage and finally to the fine reconstruction stage. Furthermore, through our point cloud-to-voxel grid shader, we transformed the fine reconstruction result from point cloud form into voxel grid (Fig. \ref{fig3}(a)). More detailed visualizations of the intermediate iterative process can be found in Supplementary Fig. 5.

\subsection{3D PA image reconstruction of in vivo animal studies}\label{subsec2_2}

Then, we performed image reconstruction of real mouse and rat experimental data using both the UBP and the SlingBAG algorithms. The in vivo animal study data, including mouse brain, rat kidney and rat liver, was acquired by Kim's lab using a hemispherical ultrasound (US) transducer array with 1024 elements and a radius of 60 mm~\cite{choi2023deep}. Each US transducer element in the array had an average center frequency of 2.02 MHz and a bandwidth of 54\%. The effective field of view (FOV) was $12.8 \text{ mm} \times 12.8 \text{ mm} \times 12.8 \text{ mm}$, and the spatial resolutions of approximately 380 µm were nearly isotropic in all directions when all 1024 US transducer elements were used~\cite{choi2023deep}. More details regarding the 3D PAI system and animal experiments can be found in literatures~\cite{choi2023deep, choi2023recent, kim20243d, kim2022deep, Kim2024inpress}.

\begin{figure}[ht]
\centering
\includegraphics[width=0.9\textwidth]{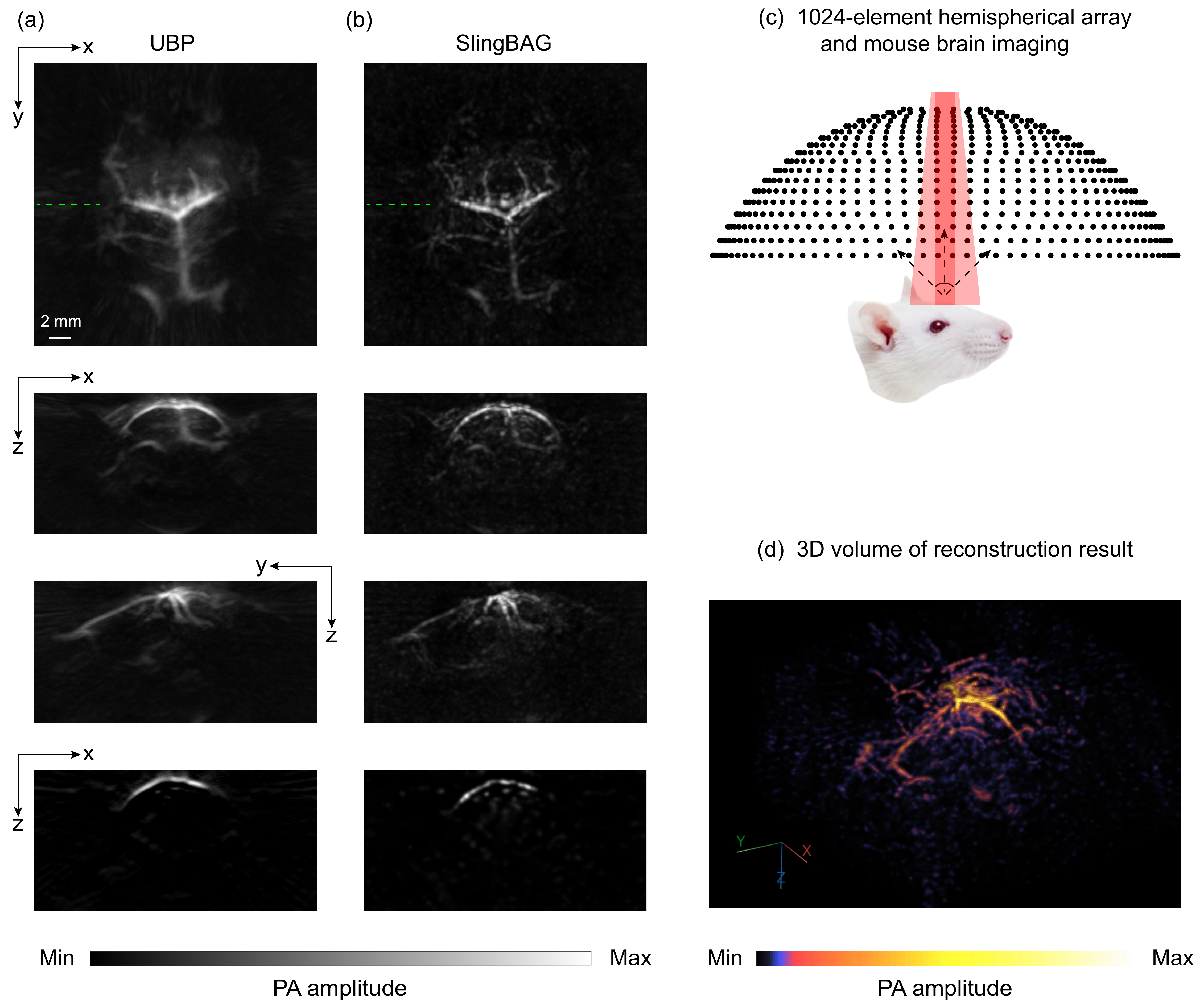}
\caption{3D PA reconstruction of a mouse brain. (a) XY Plane-MAP, XZ Plane-MAP, YZ Plane-MAP and the cross-section slice at green dashed line marked in XY Plane-MAP of the UBP 3D reconstruction results using 1024 sensor signals. (b) XY Plane-MAP, XZ Plane-MAP, YZ Plane-MAP and the cross-section slice at green dashed line marked in XY Plane-MAP of the SlingBAG 3D reconstruction results using 1024 sensor signals. (Scale: 2 mm.) (c) Schematic diagram of the hemispherical array imaging the mouse brain. (d) SlingBAG reconstruction result.}\label{fig4}
\end{figure}

\begin{figure}[H]
\centering
\includegraphics[width=1.0\textwidth]{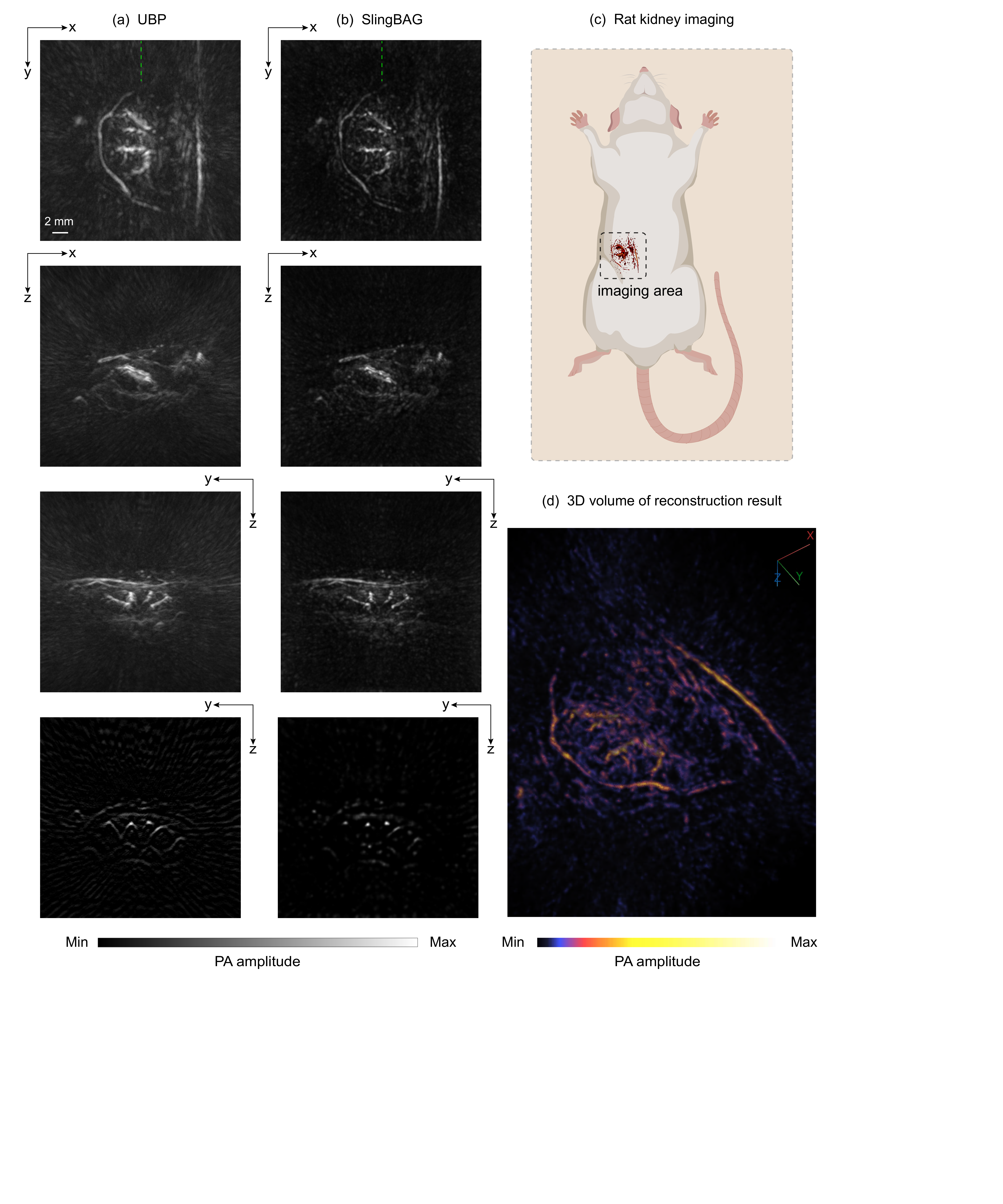}
\caption{3D PA reconstruction results of a rat kidney. (a) XY Plane-MAP, XZ Plane-MAP, YZ Plane-MAP and the cross-section slice at green dashed line marked in XY Plane-MAP of the UBP 3D reconstruction results using 1024 sensor signals. (b) XY Plane-MAP, XZ Plane-MAP, YZ Plane-MAP and the cross-section slice at green dashed line marked in XY Plane-MAP of the SlingBAG 3D reconstruction results using 1024 sensor signals. (Scale: 2 mm.) (c) Schematic diagram of the imaging area. (d) SlingBAG reconstruction result.}\label{fig5}
\end{figure}

\begin{figure}[H]
\centering
\includegraphics[width=1.0\textwidth]{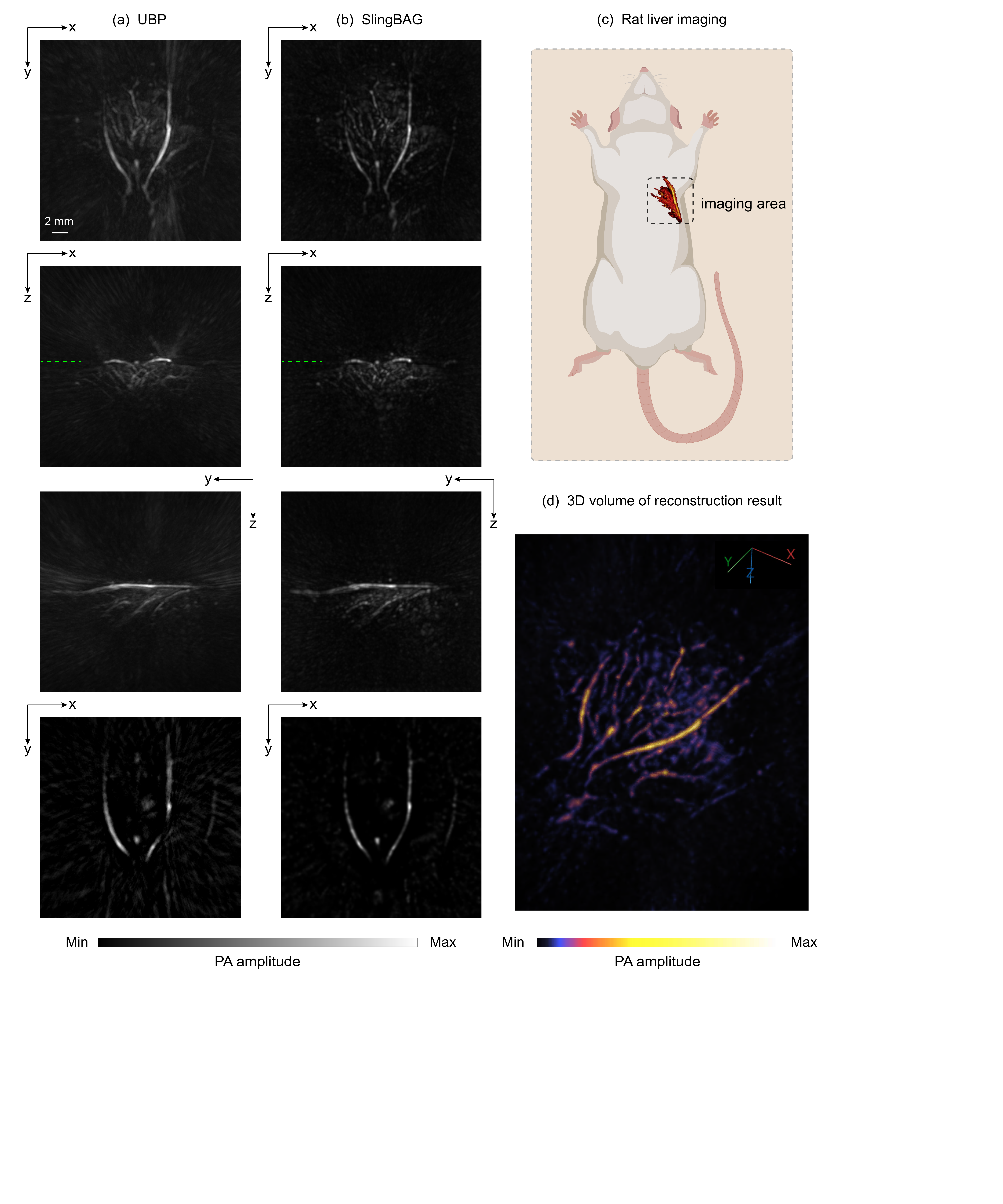}
\caption{3D PA reconstruction results of a rat liver. (a) XY Plane-MAP, XZ Plane-MAP, YZ Plane-MAP and the cross-section slice at green dashed line marked in XZ Plane-MAP of the UBP 3D reconstruction results using 1024 sensor signals. (b) XY Plane-MAP, XZ Plane-MAP, YZ Plane-MAP and the cross-section slice at green dashed line marked in XZ Plane-MAP of the SlingBAG 3D reconstruction results using 1024 sensor signals. (Scale: 2 mm.) (c) Schematic diagram of the imaging area. (d) SlingBAG reconstruction result.}\label{fig6}
\end{figure}

First, we compared the UBP and SlingBAG reconstruction results for a mouse brain using 1024 array elements (Fig. \ref{fig4}). In Fig. \ref{fig4}(a), from top to bottom, the images show the MAP results in XY Plane, XZ Plane, and YZ Plane, respectively, as well as a slice of the UBP reconstruction. Fig. \ref{fig4}(b) presents counterpart results by the SlingBAG algorithms. The 3D SlingBAG reconstruction result in Fig. \ref{fig4}(d) is available online (Supplementary Video 4). According to results, the SlingBAG reconstruction exhibits minimal artifacts and reveals far more vascular details compared with UBP results. While in the UBP result, due to artifacts induced by sparse sensors, the vascular cross-sections are almost indiscernible. Further analysis showed that the contrast-to-noise ratio (CNR) of the UBP reconstruction was 10.60, whereas the CNR of the SlingBAG reconstruction reached 61.55. We also demonstrated the SlingBAG algorithm's ability for the PA image reconstruction of in vivo mouse brain under both limited-view and sparse-view conditions in Supplementary Fig. 2-4.

Next, in Fig. \ref{fig5} and Fig. \ref{fig6}, we presented the reconstruction results of a rat kidney and a rat liver, respectively. In the cross-section slice of the reconstructed kidney image in Fig. \ref{fig5}(a), the UBP results show severe arc-shaped artifacts caused by sparse sensor configuration and limited view issues, which significantly impair the ability to resolve the fine vascular structures. In contrast, the SlingBAG algorithm effectively suppresses these arc-shaped artifacts, producing a much clearer and cleaner vascular cross-sectional image (Fig. \ref{fig5}(b)). Similarly, the SlingBAG algorithm renders the liver vasculature with much better continuity and much fewer artifacts. The 3D reconstruction results in Fig. \ref{fig5}(d) and Fig. \ref{fig6}(d) can be found in Supplementary Video 5 and Supplementary Video 6. For a quantitative comparison, the CNR values by the UBP reconstruction for the kidney and liver are 21.56 and 17.48, respectively, whereas they are 58.15 and 48.57 for SlingBAG results. These results illustrate the exceptionally high quality of image reconstruction achieved by the SlingBAG algorithm.

\subsection{Computing memory consumption analysis and comparison}\label{subsec2_3}

Besides the quality of image reconstruction, the demand of computing memory usage is also key for practical implementation. Here, we compared the computing memory consumption of the SlingBAG algorithm with that of traditional IR methods under the same experimental setup.

For the PA imaging of hand vessels using a planar matrix array with 4,900 sensors, the forward simulation combined with the backward gradient propagation of the SlingBAG algorithm required only 3.56 GB of GPU memory. In comparison, the theoretical memory consumption of traditional IR methods, such as k-Wave and the Fast Iterative Shrinkage-Thresholding Algorithm (FISTA)~\cite{beck2009fast}, is estimated to be as high as 19.32 GB, which exceeds the memory capacity limits of many consumer-grade GPUs. Similarly, for the in vivo animal imaging experiment using a hemispherical array, where the sensors are relatively far away from the reconstructed region, the SlingBAG algorithm required 2.99 GB of GPU memory, while the memory usage of traditional IR methods is estimated to be an astonishing 85.87 GB, making it virtually impractical for most accessible consumer-level GPU resources. More details of theoretical analysis for memory demand by traditional IR methods are provided in the Supplementary Note 4.

\section{Discussion}\label{sec3}

Real-time large-scale 3D PA imaging using sparsely distributed detectors has garnered significant attention. However, traditional IR reconstruction methods face great challenges due to their extremely high demand for GPU memory. This work proposes SlingBAG, a point cloud-based iterative 3D PAI reconstruction algorithm, which substantially increases speed with much less memory consumption. We have demonstrated its superior performance in achieving high-quality 3D PA images under extremely sparse sensor distributions using both simulated data and in vivo experimental data. With a feasibly accessible custom-level GPU resource, the SlingBAG algorithm can effectively suppress the reconstruction artifacts to provide high-quality 3D PA imaging.

Compared with traditional IR methods, the SlingBAG reconstruction method has two key advantages: much less memory consumption and faster reconstruction speed. For instance, the simulated hand vessels data spans dimensions of $40\text{ mm} \times 104\text{ mm} \times 80\text{ mm}$ in physical space. In traditional voxel grids, using a 0.2 mm grid interval results in a very large grid size of $200 \times 520 \times 400$ ($\sim$ $4 \times 10^7$). With this scale of grid, using 196 sensors at a 40 MHz sampling frequency (each channel requires 4096 sampling points), the data volume size makes it extremely difficult for traditional voxel-based IR methods to handle in terms of memory usage and time consumption. However, the SlingBAG method only requires less than 3 GB of GPU memory. Therefore, although traditional IR methods with powerful tools like k-Wave have shown superior performance in 2D PA imaging, they face great challenges for large-scale 3D imaging since the computing memory usage increases exponentially. Additionally, in iterative algorithms, this imposes a significant memory burden on gradient computation and optimizers, which becomes unacceptable for most research laboratories.

Moreover, concerning computational time consumption, k-Wave, as well as other tools that numerically solve PA wave function in time domain, requires computing the acoustic field at each time step over entire spatial grids, which is extremely time-consuming for large-scale 3D PA imaging. However, the SlingBAG directly calculates the far-field PA wave for the forward model using the differentiable rapid radiator. For instance, on one RTX 3090 Ti GPU, in the mentioned hand vascular case with 196 sensors and 4096 temporal sampling points, the SlingBAG takes only 15 seconds with a point cloud with 600,000 Gaussian-distributed spherical sources to finish the calculation of the forward model. In contrast, with the same computing platform, k-Wave takes more than 1000 seconds, over 55 times slower than SlingBAG. It takes less than 4 hours for SlingBAG to complete 3D PA image reconstruction with 196 sensors, while the memory overflows for k-Wave at this scale. The significant improvement in both reconstruction range and speed is largely attributable to the adaptive point cloud representation, which inherently aligns better with the sparsity characteristics in PA sources. Considering both physical model accuracy and computational cost, the point cloud representation provides a more efficient description of PA sources than voxel representation. Additionally, our differentiable rapid radiator ensures that both the forward simulation and the backward gradient propagation are fully differentiable, enabling rapid parameter updates and convergence.

The SlingBAG demonstrates high reconstruction precision. Comparing the reconstructed results to the ground truth presents its excellent quantitative accuracy in reconstructing PA sources’ both locations and pressure values. The 3D PAI results from SlingBAG can also be used to generate extra ``artificial" temporal signals for sensors at new positions with high accuracy (Supplementary Fig. 1), which is very useful for sensor interpolation research for PAI. The high reconstruction accuracy is attributable to that the point cloud model allows the source to converge to finer spatial locations, because the optimization of all Gaussian balls' positions is spatially continuous. Furthermore, the computational steps are completely differentiable, facilitating easy gradient propagation and parameter updates.

It is worth noting that the multi-order spherical discretization strategy employed in our differentiable rapid radiator is not the sole option; rather, it represents a trade-off between the required accuracy of our study and the associated memory and computational costs. The parameters for the multi-order spherical discretization are determined by Gaussian distribution in statistics (explained in Methods), which could be extended to higher orders to enhance the model's accuracy, albeit at the cost of increased computational demand.

It is important to point out that the learning rate during the iterative process has a significant impact on the model's convergence speed and reconstruction quality. Based on our experiments, it's  appropriate to set the learning rates for peak initial pressure and standard deviation to 0.05 and $4 \times 10^{-5}$, respectively, during the coarse reconstruction phase. While in the fine reconstruction phase, the suggested learning rates for the 3D coordinates, peak initial pressure, and standard deviation are $4 \times 10^{-6}$, $4 \times 10^{-3}$, and $4 \times 10^{-6}$, respectively. Interestingly, during the real mouse brain reconstruction process, we observed that although further reducing the learning rate could decrease the loss, the overall image quality becomes deteriorated instead, with more noise introduced. This phenomenon occurs because, under excessively small learning rates, the model tends to reconstruct the noise present in the original signal, resulting in noisier reconstructions.

Despite those mentioned unique advantages of SlingBAG, it is currently only valid in an acoustically homogeneous, non-viscous medium, i.e., with uniform sound speed and no attenuation due to viscosity. While traditional IR methods based on directly solving wave functions in more complex media, like k-Wave, can consider these effects. In future work, we will first tackle the non-uniform sound speed issue. One potential way is to use multi-order spherical harmonic (SH) functions to represent the anisotropic distribution of sound speeds. Specifically, for each Gaussian-distributed spherical source, in addition to peak value, standard deviation, and mean, coefficients of the SH function that represent the orientation-dependent average sound speed will be introduced. This orientation-dependent average sound speed will be computed using SH to model the acoustic radiation in various azimuth and elevation directions. The SH coefficients will be iteratively optimized, at the expense of more computational load.

Inherited from its unique IR-based superior characteristics, the SlingBAG algorithm is highly versatile and can be easily applied to a variety of array configurations, such as planar arrays, spherical arrays, as well as many more customized geometric setups, making it a powerful tool for 3D PAI reconstruction. Owing to exceptional imaging quality, reduced demand for computational resources, and significant reduction in system cost — enabled by the use of extremely sparse arrays, the SlingBAG algorithm will make a great contribution to fostering large-scale 3D PAI implementations, such as real-time whole-body small animal imaging, large-scale peripheral vascular imaging, and 3D PA imaging of breast or thyroid tumors.

\section{Methods}\label{sec4}

\subsection{Differentiable rapid radiator based on multi-order spherical discretization of Gaussian-distributed spherical PA source}\label{subsec4_1}

In the field of computer graphics, rendering describes the forward process of converting the shape, color, and material of a 3D scene into a 2D image. Conversely, reconstruction is the inverse process that restores 3D scene shape, color, and material from 2D images. Differentiable rendering, characterized by its differentiable rendering process, enables the computation of image gradients and the optimization of 3D scene parameters through methods like gradient descent, further facilitating the restoration of 3D scenes. Differentiable rendering is foundational for high-quality 3D scene rapid reconstruction technologies such as NeRF and 3D Gaussian splatting. The rendering equation used in light rendering simulates the propagation of light within a scene based on edge sampling Monte Carlo path-tracing methods. However, the acoustic radiation process in PAI differs significantly.

Corresponding to 3D vision, the forward process in PA imaging describes the propagation of ultrasound waves from PA sources with initial PA pressure to the transducer's reception, resulting in temporal acoustic signals. The inverse process reconstructs the initial PA pressure at the acoustic source from these received temporal signals.

K-Wave~\cite{treeby2010k} is a widely used simulation toolkit in the PA imaging field. It numerically solves the wave equation using the pseudo-spectral method to obtain temporal pressure signals at all grid's spatial positions from the initial PA sources. This method provides high simulation accuracy and supports extensive application scenarios by allowing settings for various medium acoustic properties, including heterogeneous sound velocity distributions. However, this method requires substantial storage space to cover all spatial grids from the source to the remote sensors. Although it works well for 2D PAI, it faces severe challenges to apply this simulation toolkit in large-scale 3D reconstruction methods due to the enormous memory consumption. Additionally, the forward propagation process using the pseudo-spectral method does not support direct gradient back-propagation, limiting its performance in IR methods.

\begin{figure}[ht]
\centering
\includegraphics[width=1.0\textwidth]{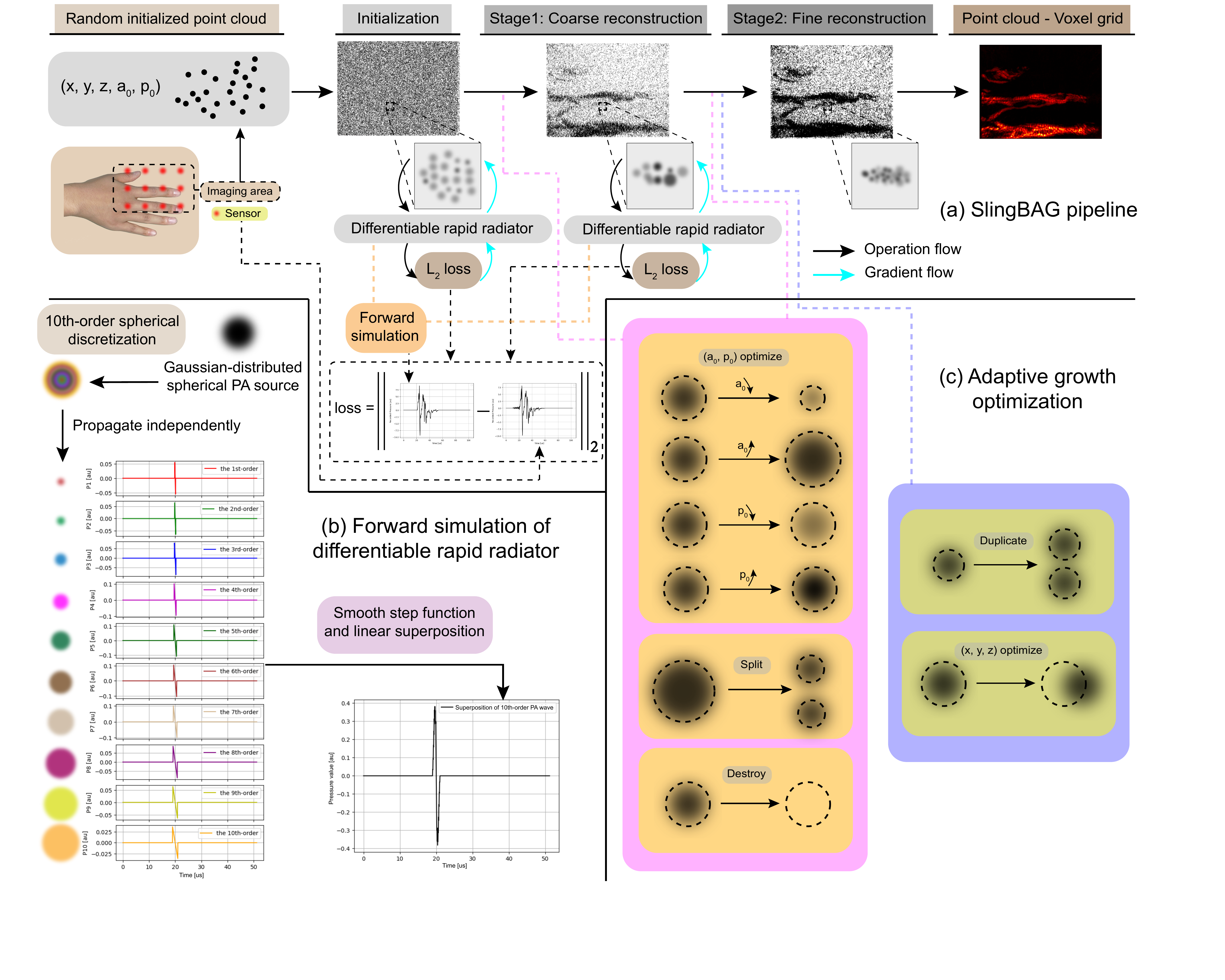}
\caption{The overview framework of Sliding Gaussian Ball Adaptive Growth algorithm. (a) The SlingBAG pipeline. (b) The forward simulation of differentiable rapid radiator. (c) Adaptive growth optimization based on iterative point cloud.}\label{fig:pipeline_gaussian}
\end{figure}

Inspired by differentiable rendering techniques from computer graphics, and based on the special analytical solution from the PA wave propagation, we propose the concept of ``differentiable acoustic radiation" and introduce a unique differentiable rapid radiator for PA imaging. The entire process is fully differentiable and friendly to parallel computing.

\begin{figure}[H]
\centering
\includegraphics[width=1.0\textwidth]{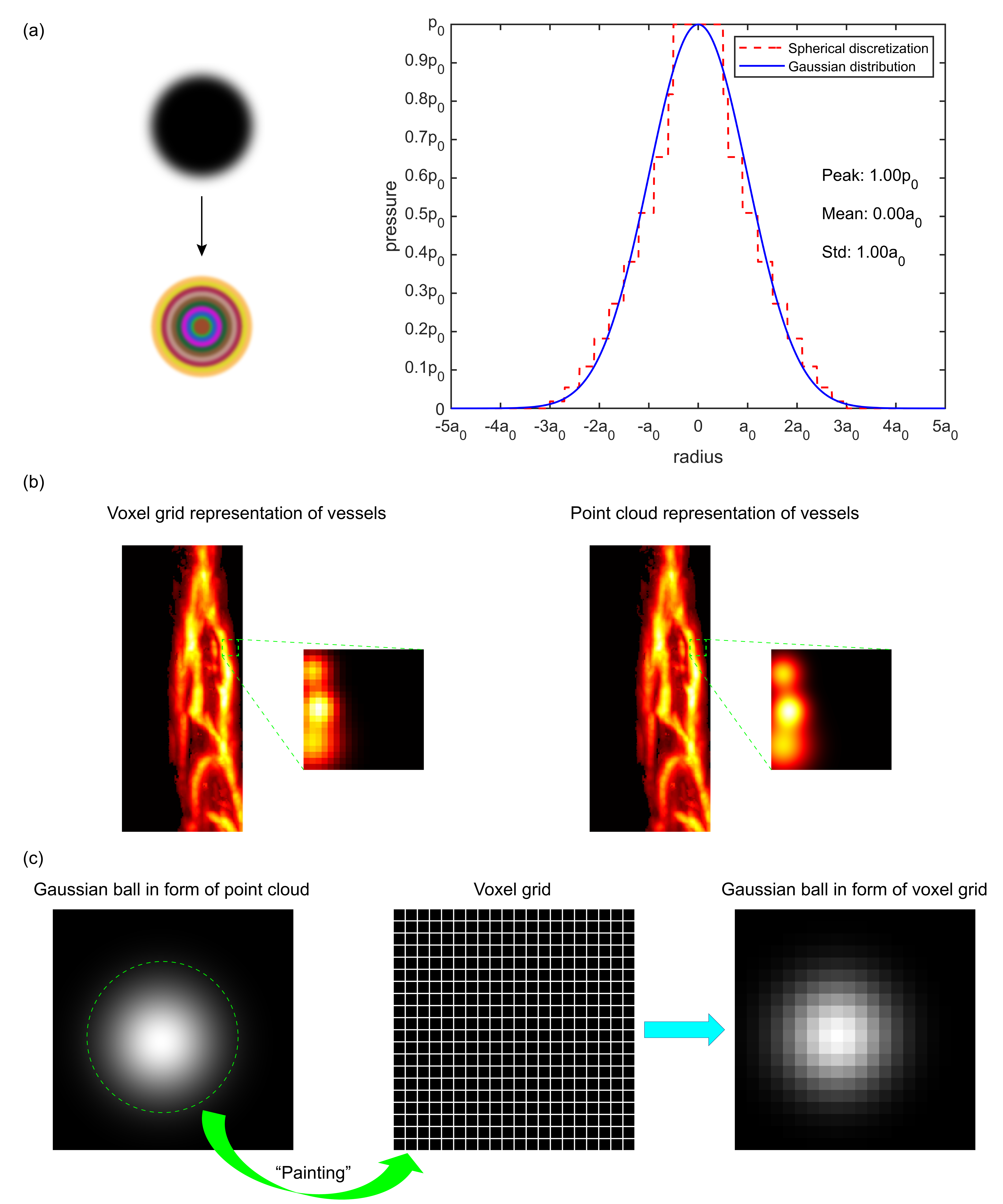}
\caption{The discretization, representation and conversion of point cloud based Gaussian balls. (a) The 10th-order spherical discretization of Gaussian-distributed Spherical PA source. (b) Comparison of voxel grid representation and point cloud-based Gaussian ball representation of a PA scene. (c) The conversion of Gaussian ball from the form of point cloud to the form of voxel grid.}\label{fig:method}
\end{figure}

As shown in Fig. \ref{fig:method}(b), this model uses a series of isotropic Gaussian-distributed PA sources to explicitly represent 3D PA scenes, with each Gaussian PA source characterized by three attributes: its peak initial pressure $p_0$, standard deviation $a_0$, and 3D coordinates $(x_s, y_s, z_s)$ (i.e., the mean of the spherical Gaussian distribution). These attributes are stored in the form of point cloud and iteratively updated during the reconstruction process. The entire PA propagation is viewed as a superposition of the independent PA wave propagation of all Gaussian PA sources within the scene. Each Gaussian PA source is a combination of multi-order concentric homogeneous spheres, which follows the three-sigma rule of Gaussian distribution, as indicated in Fig. \ref{fig:method}(a). For each sphere, its initial pressure and radius are computed based on the peak initial pressure $p_0$, standard deviation $a_0$, combined with fixed weights — a process referred to as multi-order spherical discretization (Fig. \ref{fig:pipeline_gaussian}(b)). The reason to choose spherical shape is that the propagation equation of a homogeneous spherical PA source has an analytical solution, which helps to build a differentiable rapid radiator. The details of the forward simulation and the backward propagation of the differentiable rapid radiator are shown as below.

\textbf{(i) Forward simulation.} For a Gaussian PA source with peak initial pressure $p_0$, standard deviation $a_0$, and 3D coordinates $(x_s, y_s, z_s)$, we detail parameters of its 10th-order concentric spherical discretization: the common center of all 10 spheres is located at $(x_s, y_s, z_s)$, and their radii and initial PA pressures distributed on each sphere are as follows in Tab. \ref{tb:tb1}:

\begin{table}[ht]
\centering
\caption{Radii and pressures of 10 spheres}
\begin{tabular}{cccccccccccc}
\toprule
Radius & $0.5a_0$ & $0.6a_0$ & $0.9a_0$ & $1.2a_0$ & $1.5a_0$ & $1.8a_0$ & $2.1a_0$ & $2.4a_0$ & $2.7a_0$ & $3.0a_0$ \\
\midrule
Pressure & $\frac{10}{55}p_0$ & $\frac{9}{55}p_0$ & $\frac{8}{55}p_0$ & $\frac{7}{55}p_0$ & $\frac{6}{55}p_0$ & $\frac{5}{55}p_0$ & $\frac{4}{55}p_0$ & $\frac{3}{55}p_0$ & $\frac{2}{55}p_0$ & $\frac{1}{55}p_0$ \\
\bottomrule
\end{tabular}
\label{tb:tb1}
\end{table}

It is worth noting that this discretization method is not unique. The choice of the 10th-order balances model accuracy and computational complexity. Using higher-order spherical discretization can further improve computational accuracy, but the computational load will increase linearly with the order.

The analytical solution for the PA wave propagation of a uniform sphere PA source excited by a Dirac delta laser pulse~\cite{xu2005universal} is:

\begin{equation}
\begin{split}
    p(R, t) &= p_0\frac{R - v_s t}{2R}\left(u(R - v_s t + a_0) - u(R - v_s t - a_0)\right),\\
    \quad \\
    \text{where}\quad R&=\sqrt{(x_r-x_s)^2+(y_r-y_s)+(z_r-z_s)^2},\\
    \quad \\
    u(x) &= \left\{ \begin{array}{ll} 0, & x < 0 \\
    \quad \\
    1, & x \geq 0 \end{array}\right.,\quad \delta(x)=\frac{\mathrm{d}}{\mathrm{d}x}u(x),
    \label{eq:eq1}
\end{split}
\end{equation}

where $(x_s,y_s,z_s)$ is the position of the sphere’s center, $v_s$ is the speed of sound, $t$ is the propagation time, $p_0$ is the initial acoustic pressure of the sphere, $a_0$ is the radius of the sphere, $(x_r,y_r,z_r)$ is the sensor coordinate, $R$ is the Euclidean distance from the sensor to the center of the sphere (thus $R>a_0$), and $u(x)$ is the unit step function.

Based on the analytical solution and superposition of waves from all concentric spheres, we can derive the 10th-order spherical acoustic pressure propagation expression $P_{\mathrm{10ball}}$. The received acoustic pressure signal at any spatial location is obtained by the linear superposition of the acoustic pressure propagation expressions from all Gaussian spherical sources.

Due to the presence of the unit step function $u(x)$, the 10th-order spherical discretization is not differentiable yet. To address this, we constructed a new formula using the differentiable error function $\mathrm{erf}(x)$ to approximate $u(x)$ with high precision:

\begin{equation}
\begin{split}
    u_\epsilon(x)&=\frac{1}{2}(1+\mathrm{erf}(\epsilon x)),\\
    \quad \\
    \text{where}\quad\mathrm{erf}(x)&=\frac{2}{\sqrt{\pi }}\int _0^x{e}^{-{t}^{2}}\mathrm{d}t.
    \label{eq:eq2}
\end{split}
\end{equation}

The value of factor $\epsilon$ is related to the minimum $a_0$ (representing the spatial resolution) in the SlingBAG algorithm. In this study, we set the factor $\epsilon$ to $10^{-6}$/µm, and $u_\epsilon(x)$ increases from 0.0002 to 0.9998 as $x$ increases from -2.5 µm ($-2.5 \times 10^{-6}$) to 2.5 µm ($2.5 \times 10^{-6}$) near zero. Therefore, $u_\epsilon(x)$ is a high-precision approximation of the step function for $a_0$ ($>$ 50 µm), valid for most 3D PAI applications for deep tissue imaging.

Thus, we obtain the final acoustic pressure propagation expression for a Gaussian spherical source with peak initial pressure $p_0$, standard deviation $a_0$ and spatial coordinates $(x_s, y_s, z_s)$ based on the 10th-order spherical discretization, which is also fully differentiable:

\begin{equation}
\begin{split}
    p(R, t) &= \sum_{i=1}^{10}p_0^i\frac{D}{2R}\left(u_\epsilon(D + a_0^i) - u_\epsilon(D - a_0^i)\right),\\
    \quad \\
    \text{where}\quad D &= R - v_s t,
    \label{eq:eq3}
\end{split}
\end{equation}

where $a_0^i$ and $p_0^i$ are the radius and initial pressure of the $i$th-order sphere in Tab. \ref{tb:tb1}.

\textbf{(ii) Backward propagation.} After using a differentiable modified step function, our acoustic pressure propagation function $p(R,t)$ becomes fully differentiable, and the derivative of the modified step function in Eq.(\ref{eq:eq2}) is:

\begin{equation}
    \delta_\epsilon(x)=\frac{\mathrm{d}}{\mathrm{d}x}u_\epsilon(x)=\frac{\epsilon}{\sqrt{\pi}}\exp\left(-{\epsilon}^2x^2\right),
    \label{eq:eq4}
\end{equation}

which is a good approximation of the Dirac delta function $\delta(x)$.

Then, in the end-to-end training process, we need to compute the derivatives of $p(R,t)$ with respect to the peak initial pressure $p_0$, standard deviation $a_0$, and the spatial center coordinates $(x_s,y_s,z_s)$ to propagate the gradient backward. Taking the partial derivative of Eq.(\ref{eq:eq3}), we can get:

\begin{equation}
\begin{split}
    \frac{\partial p}{\partial p_0}&=\sum_{i=1}^{10}\frac{p_0^i}{p_0}\frac{D}{2R}\left(u_\epsilon(D + a_0^i) - u_\epsilon(D - a_0^i)\right), \\
    \quad \\
    \frac{\partial p}{\partial a_0}&=\sum_{i=1}^{10}\frac{a_0^i}{a_0}p_0^i\frac{D}{2R}\left(\delta_\epsilon(D + a_0^i) + \delta_\epsilon(D - a_0^i)\right), \\
    \quad \\
    \frac{\partial p}{\partial x_s}&=\left(\frac{\partial p}{\partial R}+\frac{\partial p}{\partial D}\frac{\partial D}{\partial R}\right)\frac{\partial R}{\partial x_s} \\
    &=-\frac{x_r-x_s}{R}\sum_{i=1}^{10}p_0^i\bigg(\frac{v_s t}{2R^2}\left(u_\epsilon(D + a_0^i) - u_\epsilon(D - a_0^i)\right) \\
    &\qquad+\frac{D}{2R}\left(\delta_\epsilon(D + a_0^i) - \delta_\epsilon(D - a_0^i)\right)\bigg).
    \label{eq:eq5}
\end{split}
\end{equation}

The derivatives with respect to $y_s$ and $z_s$ are similar to the derivative with respect to $x_s$ shown above. The above expression demonstrates the back-propagation computation of our differentiable rapid radiator. Both forward and backward models are implemented with GPU-accelerated code based on Taichi~\cite{hu2019taichi} and pure CUDA syntax.

\subsection{Sliding Gaussian ball adaptive growth point cloud model}\label{subsec4_2}

The overview framework of the SlingBAG algorithm is shown in Fig. \ref{fig:pipeline_gaussian}(a). We start by randomly initializing the point cloud within the field of view. All attributes of each Gaussian ball (10th-order concentric spheres) are randomly sampled under a uniform distribution within a certain range. This randomly initialized point cloud serves as the initial PA source. PA signals received by each sensor are calculated by the differentiable rapid radiator. By calculating the loss between these simulated signals and the actual sensor data and performing gradient back-propagation, the model then updates all attributes of the points in the point cloud. All the optimization methods of each point in the point cloud are demonstrated in Fig. \ref{fig:pipeline_gaussian}(c).

It is important to emphasize that the IR reconstruction contains two successive reconstruction stages: coarse reconstruction stage and fine reconstruction stage. In the first stage, only the standard deviation $a_0$ and peak initial acoustic pressures $p_0$ are updated, while the spatial coordinates of the points remain unchanged, and the model's duplication feature is not activated. This stage helps rapid convergence while maintaining computational efficiency.

\begin{algorithm}
    \caption{SlingBAG}
    \label{alg:SlingBAG}
    \renewcommand{\algorithmicrequire}{\textbf{Input:}}
    \renewcommand{\algorithmicensure}{\textbf{Output:}}
    \begin{algorithmic}[1]
        \Require $Z_0$: Randomly initialized point cloud
        
        $S_0$: Real signal
        
        $\left(x,y,z,a_0,p_0\right)$: Attributes of each point
        \Ensure 3D reconstruction result in form of voxel grid
        \State $Z_1=Z_0$
        \While{\textit{not converged}}\Comment{Coarse reconstruction stage}
            \State Generate signal $S_1$ by forward simulation with differentiable rapid radiator
            \State Calculate loss: $L=\|S_1-S_0\|_2$
            \State Gradient back-propagate by differentiable rapid radiator: $\frac{\partial L}{\partial a_0},\frac{\partial L}{\partial p_0}$
            \State Update on $\left(a_0,p_0\right)$
            \State Perform splitting and destroying of points in $Z_1$
        \EndWhile
        \State $Z_2=Z_1$
        \While{\textit{not converged}}\Comment{Fine reconstruction stage}
            \State Generate signal $S_2$ by forward simulation with differentiable rapid radiator
            \State Calculate loss: $L=\|S_2-S_0\|_2$
            \State Gradient back-propagate by differentiable rapid radiator: $\frac{\partial L}{\partial x},\frac{\partial L}{\partial y},\frac{\partial L}{\partial z},\frac{\partial L}{\partial a_0},\frac{\partial L}{\partial p_0}$
            \State Update on $\left(x,y,z,a_0,p_0\right)$
            \State Perform duplicating, splitting and destroying of points in $Z_2$
        \EndWhile
        \For{\textit{each point} \textbf{in} $Z_2$}\Comment{Point cloud to voxel grid}
            \State Convert into voxel grid through point cloud-voxel grid shader
        \EndFor
    \end{algorithmic}
\end{algorithm}

In both coarse reconstruction stage and fine reconstruction stage, after a certain number of iterations, the model undergoes adaptive density optimization of the point cloud to ensure computational efficiency and prevent convergence to sub-optimal solutions that are physically unreal. Points with excessively low peak initial acoustic pressure $p_0$ or standard deviation $a_0$ are discarded to maximize resource savings and expedite iteration speed. For points with excessively large $a_0$, the model performs a splitting operation. This involves splitting an initially large sphere into two smaller spheres, each with half the original radius, while maintaining the same peak initial acoustic pressure $p_0$. The spatially splitting direction is aligned with the gradient direction of the point position.

As the loss begins to converge, the model further enhances its capabilities by incorporating position coordinate updates into the optimizer alongside updates for peak initial acoustic pressure and standard deviations, and the point cloud enters into the fine reconstruction stage. The model then activates its duplication feature at specific iteration counts, enabling all points in the current point cloud to be duplicated along their respective position gradients. This approach increases the point cloud's density and allows for more precise control and fine-tuning of each point, leading to more detailed and accurate 3D reconstructions.

Through these processes, the model can further refine the point cloud, improving the contrast and detail of the 3D reconstruction. This adaptive and iterative approach facilitates high-quality 3D PA imaging reconstructions even under conditions of extremely sparse sensor distributions.

\subsection{Physically-based point cloud to voxel grid shader}\label{subsec4_3}

While the 3D point cloud resulting from iterative processing encapsulates all the information about the reconstructed PA sources, directly visualizing it does not effectively convey each point's standard deviation and peak initial acoustic pressure. Since voxel grids are widely used for 3D visualization of PA imaging results, we then continue to develop a converter, termed the ``Point Cloud to Voxel Grid Shader", which transforms point cloud data into voxel grid information, as shown in Fig. \ref{fig:method}(c).

Due to the density of the point cloud, its coordinates may not be perfectly aligned with the voxel grid nodes. The shader first aligns the voxel grid coordinate space with the point cloud coordinate space in three dimensions. Next, it iterates over all the points in the point cloud, reading each point's peak initial acoustic pressure and standard deviation, which correspond to the gray-scale intensity and brush size, respectively.

Here’s a detailed breakdown of the conversion process:

\begin{enumerate}
\item Alignment of Coordinate Spaces: The voxel grid coordinate space and the point cloud coordinate space are aligned in 3D space to ensure accurate correspondence.

\item Reading Point Attributes: For each point in the point cloud, the shader reads the peak initial acoustic pressure and standard deviation information. These attributes, representing gray-scale intensity and brush size, are adjusted based on the parameters from the 10th-order spherical discretization.

\item Voxel Grid Coloring: The shader ``paints" each point into the 3D space using the adjusted gray-scale intensity and brush size. The markings interact with the voxel grid, mapping the gray-scale intensity onto the grid nodes. Intensity values at the voxel grid nodes are accumulated when multiple brushstrokes overlap, ensuring a complete rendering of the voxel grid from the point cloud data.
\end{enumerate}

By converting point cloud data into a voxel grid using this physically-based point cloud to voxel grid shader, we achieve comprehensive and detailed 3D visualization that accurately reflects the initial acoustic pressures and resolutions from the original point cloud data. This method leverages the strengths of both point cloud and voxel grid representations, enhancing visualization for analysis and interpretation in the field of PA imaging.

\backmatter

\section*{Acknowledgements}

This research was supported by the following grants: the National Key R\&D Program of China (No. 2023YFC2411700, No. 2017YFE0104200); the Beijing Natural Science Foundation (No. 7232177); the National Natural Science Foundation of China (62441204, 62472213); the Basic Science Research Program through the National Research Foundation of Korea (NRF) funded by the Ministry of Education (2020R1A6A1A03047902).

We would like to express our gratitude to Professor Chao Tian and Mr. Zhijian Tan from the University of Science and Technology of China for their selfless support.

\section*{Declarations}

\subsection*{Funding}

The National Key R\&D Program of China (No. 2023YFC2411700, No. 2017YFE0104200); the Beijing Natural Science Foundation (No. 7232177); the National Natural Science Foundation of China (62441204, 62472213); the Basic Science Research Program through the National Research Foundation of Korea (NRF) funded by the Ministry of Education (2020R1A6A1A03047902).

\subsection*{Competing interests}

All authors declare no competing interests.

\subsection*{Data availability}

The data supporting the findings of this study are provided within the Article and its Supplementary Information. The raw and analysed datasets generated during the study are available for research purposes from the corresponding authors on reasonable request.

\subsection*{Code availability}

The source code and data for SlingBAG, along with supplementary materials and demonstration videos, are now publicly available in the following GitHub repository: \href{https://github.com/JaegerCQ/SlingBAG}{https://github.com/JaegerCQ/SlingBAG}.

\subsection*{Author contributions}

Shuang Li, Yibing Wang and Jian Gao conceived and designed the study. Chulhong Kim and Seongwook Choi provided the experimental data. Yu Zhang and Qian Chen contributed to the design of the simulation experiments. Changhui Li and Yao Yao supervised the study. All of the authors contributed to writing the paper.

\section*{Supplementary information}


Please refer to our separate Supplementary Information file for further details.

\bibliography{references}% common bib file

\end{document}